\documentclass[runningheads]{llncs}
\usepackage{graphicx}
\usepackage{mathtools}
\usepackage{todonotes}
\usepackage{hyperref}
\usepackage{amsmath,amssymb}
\usepackage{appendix}
\usepackage{arydshln}

\usepackage{subcaption}
\usepackage{mwe}

\usepackage{fixmath}
\usepackage{bm}

\DeclarePairedDelimiterX{\infdivx}[2]{[}{]}{%
  #1\;\delimsize\|\;#2%
}

\DeclareMathOperator{\E}{\mathbb{E}}
\DeclareMathOperator{\Var}{\mathrm{Var}}

\begin{document}
\title{Learning to Predict Error for MRI Reconstruction}
%
%
\author{Shi Hu\inst{1} \and Nicola Pezzotti\inst{2,3} \and Max Welling\inst{1}}
\authorrunning{S. Hu et al.}
%
\institute{University of Amsterdam \and Philips Research \and Eindhoven University of Technology}
\maketitle              
\begin{abstract}
In healthcare applications, predictive uncertainty has been used to assess predictive accuracy. In this paper, we demonstrate that predictive uncertainty estimated by the current methods does not highly correlate with prediction error by decomposing the latter into random and systematic errors, and showing that the former is equivalent to the variance of the random error. In addition, we observe that current methods unnecessarily compromise performance by modifying the model and training loss to estimate the target and uncertainty jointly. We show that estimating them separately without modifications improves performance. Following this, we propose a novel method that estimates the target labels and magnitude of the prediction error in two steps. We demonstrate this method on a large-scale MRI reconstruction task, and achieve significantly better results than the state-of-the-art uncertainty estimation methods.

\end{abstract}
\section{Introduction}

Healthcare has been increasingly facilitated by artificial intelligence technologies \cite{davenport}. Uncertainty is ubiquitous in these technologies, and it can arise due to randomness or imperfect knowledge \cite{alea_or_epi}, such as the disagreement among human annotators, missing entries in electronic health records, or occlusions in MRIs. Kennedy and O'Hagan \cite{kennedy} lists six sources of uncertainty that affect predictive outcomes, which include data noise, input variability, model structure and parameters, optimization, and interpolation. Unfortunately, the current methods in deep learning \cite{kg,deepensemble} quantify only two of them, which are data noise and model parameters; in addition, they use the sum of the two uncertainties to estimate predictive uncertainty. Lastly, they need to modify the model structure and training objective to estimate the target and uncertainty jointly. 

In this paper, we show that their estimated predictive uncertainty cannot highly correlate with prediction error; in addition, if we estimate the target and uncertainty separately, the performance improves since the model and training loss are unchanged, and regularization schemes such as early stopping can have separate effects on the two estimates. Following this, we propose a novel two-step method where we train one deep model to estimate the target, and another the magnitude of the prediction error. We demonstrate this method on a large-scale MRI reconstruction task, and achieve significantly better results than the state-of-the-art uncertainty estimation methods.

\section{Notation}
We denote an input by $x$, a noisy target by $y(x)$, and the true target by $h(x)$. The noise $\epsilon(x)$ is assumed to be additive and Gaussian, i.e. $y(x) = h(x) + \epsilon(x)$ where $\epsilon(x) \sim \mathcal{N}(0, \sigma^2(x))$, and we will refer to the noise level $\sigma(x)$ as ``sigma". Further, the estimates are marked with the caret symbol, e.g., $\hat{h}(x)$ is an estimate of $h(x)$. Unless otherwise stated, the model structure $\mathcal{M}$, training data $\mathcal{D}$ and optimizer $\mathcal{O}$ are all fixed. The expectation of any estimate is taken over the random seed $s$. For clarity of notation, we omit these symbols when possible, e.g., the expected estimate of the true target $\E_s\big[\hat{h}(x;s, \mathcal{M},\mathcal{D},\mathcal{O})\big]$ is abbreviated to $\E[\hat{h}(x)]$. 

\section{An Anatomy of Prediction Error}

To estimate the magnitude of the prediction error (or squared error) on unseen data, we first decompose the prediction error into systematic and random errors, and analyze the two separately:

\begin{align}
    \underbrace{y(x) - \hat{h}(x)}_{\text{prediction error}} = \underbrace{h(x) - \E[\hat{h}(x)]}_{\text{systematic error}} + \underbrace{\big[y(x) - h(x)\big] + \big[\E[\hat{h}(x)] - \hat{h}(x)\big]}_{\text{random error}}. \label{eq:errordecomp}
\end{align}

Systematic error is the difference between the true target and the expected estimate of the true target \cite{gum}. In deep learning, it can be reduced if we have better knowledge regarding the model structure, or can diminish the gap between the global optimum and the local optimum obtained by the optimizer. The lack of knowledge of the model structure and optimization process reflects two types of epistemic uncertainties, which are called structural and algorithmic uncertainties \cite{kennedy}. Unfortunately, to the best of our knowledge, neither has been quantified by the current deep learning methods. 

The total random error equals the prediction error minus the systematic error \cite{gum}. The first random error $y(x) - h(x) = \epsilon(x)$ represents the label noise, which is unpredictable. However, its conditional variance is equivalent to that of the noisy targets, i.e., $\sigma^2(x) = \Var[\epsilon(x)] = \Var[y(x) - h(x)] = \Var[y(x)]$. If we can access multiple noisy targets per $x$, we can predict their variance through supervised learning \cite{hu}. Otherwise, to estimate $\sigma(x)$, we need to assume that it is smooth over $x$, then predict it along with the true target \cite{nix}. Likewise, the second random error $\E[\hat{h}(x)] - \hat{h}(x)$ is also unpredictable, but its variance $\Var\big[\E[\hat{h}(x)] - \hat{h}(x)\big] = \Var[\hat{h}(x)]$ can be estimated using Monte Carlo dropout \cite{mcdropout,kg} or ensemble methods \cite{deepensemble}. Previous works \cite{mcdropout,kg} refer to the two variances as aleatoric and model (or epistemic) uncertainty. 

\begin{figure}[t]
\includegraphics[width=\textwidth]{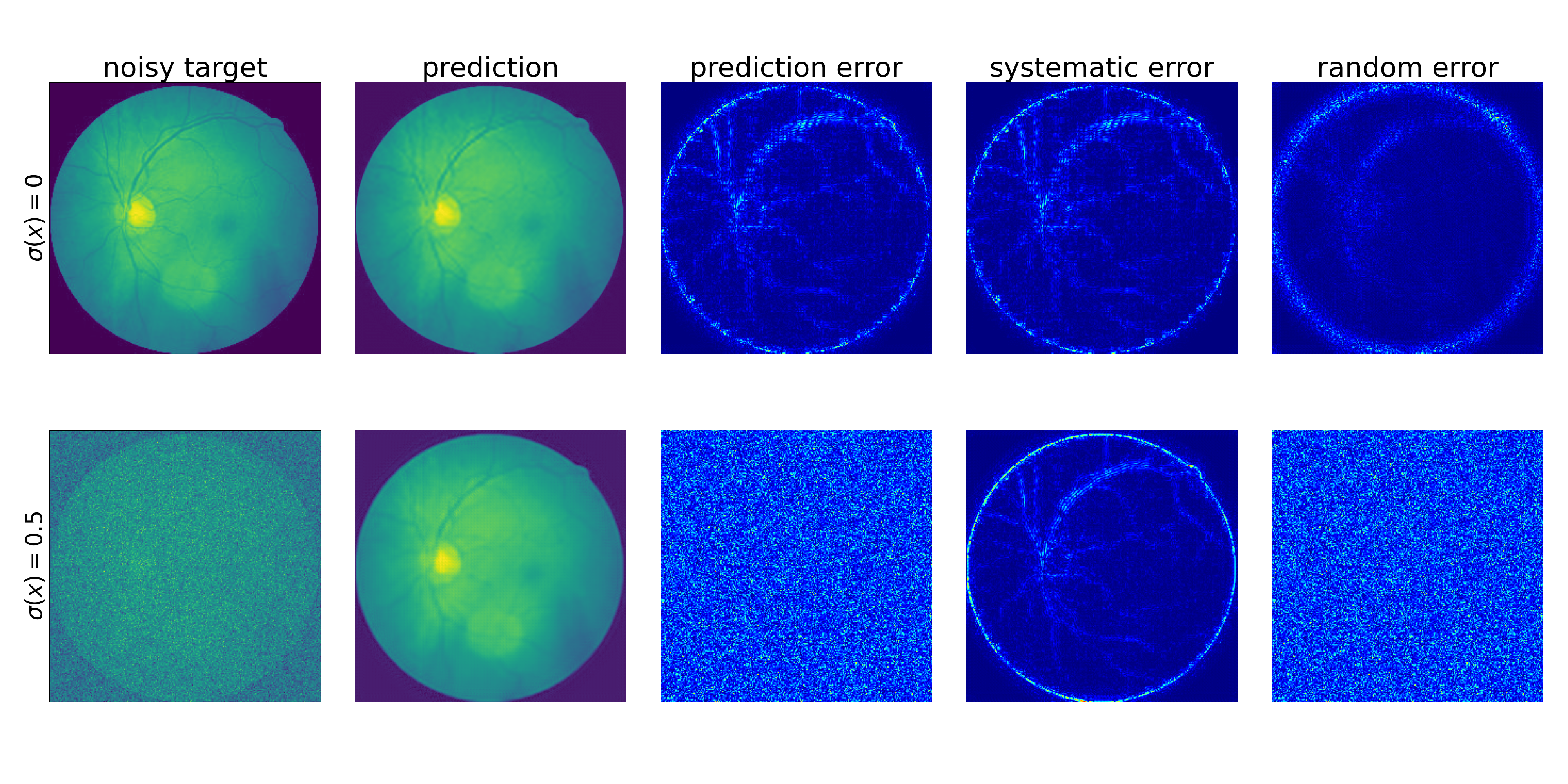}
\caption{The error decomposition for a super-resolution prediction (the retina image is shown in the luminance channel in YCbCr colour space). The target noise is part of the random error, which does not affect the systematic error.}
\label{fig:error_decomp}
\end{figure}

In summary, systematic error is deterministic, and predictable if the true target is a smooth function \cite{cybenko}. On the other hand, random error is unpredictable, so it can be best estimated using its expectation. An illustration of the error decomposition on a retina image is shown in Fig. \ref{fig:error_decomp}.

\section{What Does the Predictive Uncertainty Quantify?}

Due to the random error, the magnitude of the prediction error cannot be precisely known, so it can be best estimated with the expected squared error that integrates out the randomness in the label noise and random seed. This error can be decomposed using the bias-variance decomposition \cite{hastie} as the following:

\begin{align}
    \underbrace{\E\big[(y(x) - \hat{h}(x))^2\big]}_{\text{expected squared error}} = & \underbrace{\sigma^2(x) + \Var[\hat{h}(x)]}_{\text{variance of random error}} + \underbrace{(h(x) - \E[\hat{h}(x)])^2}_{\text{squared systematic error}}, \label{eq:bvdecomp} 
\end{align}

\noindent which means it is influenced by both random and systematic errors. However, several recent works \cite{depeweg,hu,kg,deepensemble} estimate predictive uncertainty by combining the uncertainties in data noise and model parameters under the law of total variance as follows:

\begin{align}
    \underbrace{\Var[\hat{y}(x)]}_{\text{predictive uncertainty}} = \underbrace{\E[\hat{\sigma}^2(x)]}_{\text{aleatoric uncertainty}} + \underbrace{\Var[\hat{h}(x)]}_{\text{model uncertainty}}. \label{eq:uncdecomp}
\end{align}

\noindent This predictive uncertainty is equivalent to the variance of the random error, and therefore does not contain the systematic error. As a result, it underestimates the squared error on average.

\section{A Two-Step Estimation Method}

We have shown that the current predictive uncertainty cannot highly correlate with the squared error, especially when the systematic error is high. In this section, we introduce a simple method that estimates the true target and expected squared error in two steps. We assume that for a given task, the predictive model and its training objective are known, and our method is as follows. First, we train the model to estimate only the target. After training, for each training input $x$, we compute the squared error $e^2(x) = (y(x) - \hat{h}(x))^2$ as an unbiased estimator of its expectation $\E[e^2(x)]$. Then, we train the same model from scratch to estimate $e^2(x)$, where the loss function can simply be the $L_1$ or $L_2$ loss.

This two-step method has two benefits. First, to estimate aleatoric uncertainty, the current methods \cite{kg,deepensemble} need to add a second branch to the model to output $\hat{\sigma}^2(x)$, and incorporate it into the training objective (we refer to this as a ``two-head" model). However, it is challenging to incorporate $\hat{\sigma}^2(x)$ into complex objective functions, such as the structural similarity index measure (SSIM) loss \cite{ssim} commonly used in the MRI reconstruction models \cite{nicola,irim_fastmri}. Furthermore, if the original training objective is a combination of multiple losses, it is non-trivial to include $\hat{\sigma}^2(x)$ without affecting the target prediction accuracy. In comparison, the two-step method does not need to modify the model or training loss. Second, the true target and aleatoric uncertainty are independent quantities, but the two-head model estimates them jointly using the same set of hyperparameters, including the number of epochs. Since the two estimands can have different magnitudes, there is no guarantee that they will reach their best estimates simultaneously. The two-step method can avoid this problem by applying early stopping, which prevents overfitting in each step.

\section{Experiments}

\subsection{Datasets}
\textbf{Retina} We use the diabetic retinopathy dataset\footnote{\url{https://www.kaggle.com/c/diabetic-retinopathy-detection}} for a synthetic single image super-resolution task, where we predict the high resolution (HR) image from its low resolution (LR) counterpart. We randomly sample 500 good-quality square images and resize each to $255 \times 255$ as the HR image, and use the downsampled $85 \times 85$ image as the LR counterpart (i.e., the upscaling factor is 3). We split this dataset into 200/100/200 training/validation/test images.

\noindent\textbf{FastMRI} The fastMRI dataset \cite{fastMRI} contains fully anonymized clinical MR images and raw MR measurements. We use the multi-coil knee dataset for a reconstruction task, where we predict the fully sampled MR image from its undersampled image with 4- or 8-time acceleration. The dataset contains a training, validation, test and challenge set, but only the first two provide fully sampled data, which are used for evaluation. Hence, we randomly split the validation set into a validation and test set. After the split, there are 973/59/140 training/validation/test MRI volumes.

\subsection{Single Image Super-Resolution}

We run a set of synthetic experiments on a single image super-resolution task using the retina dataset, where the HR images used as the training targets are corrupted by the pixel-wise Gaussian noise $\epsilon(x) \sim \mathcal{N}(0, \sigma^2(x))$. For ease of comparison, we use the efficient sub-pixel convolutional
neural network (ESPCN) \cite{espcn}, which is a feedforward network with the mean squared error (MSE) loss: $(y(x) - \hat{h}(x))^2$, and it processes the images in the luminance channel in YCbCr colour space. 

We compare the prediction accuracies of the true target and sigma estimates among the original, two-head, and two-head (2x) models. The first two use the same number of parameters (excluding the extra branch in the two-head), and the last uses twice as many proportional to the model structure. To implement the two-head models, we duplicate the last layer of the ESPCN and append both to the penultimate layer. The uncertainty estimate is incorporated into the MSE loss, which becomes the Gaussian negative log-likelihood (NLL) loss: $\frac{(y(x) - \hat{h}(x))^2}{\hat{\sigma}^2(x)} + \log \hat{\sigma}^2(x)$. To avoid the division by zero error, we use the numerical stable implementation as in \cite{kg}. We train 1000 epochs using the Adam optimizer \cite{adam} with learning rate $10^{-4}$, though we observe that learning rates do not affect the relative performance.

\begin{figure}[t]
\includegraphics[width=\textwidth]{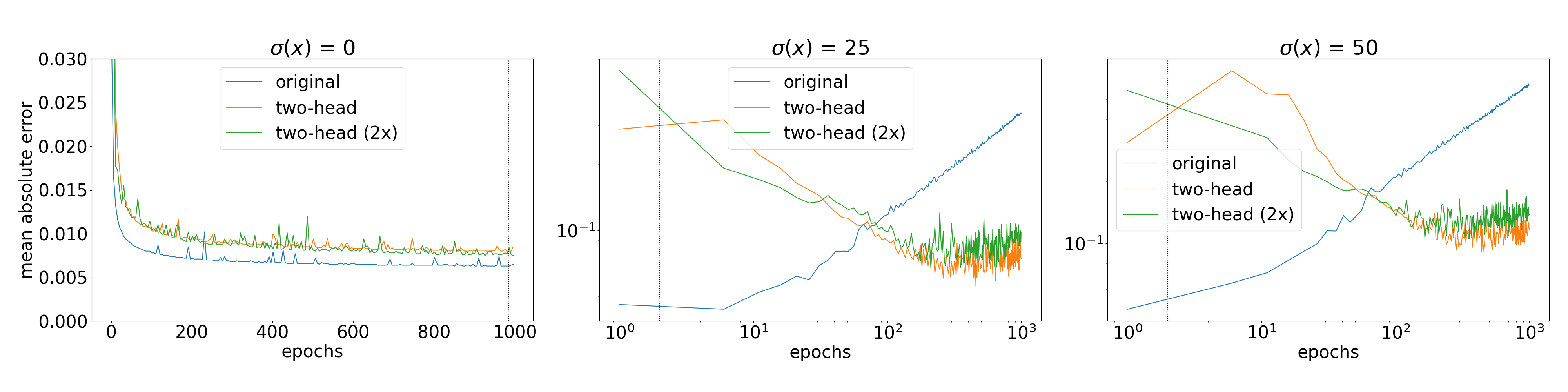}
\includegraphics[width=\textwidth]{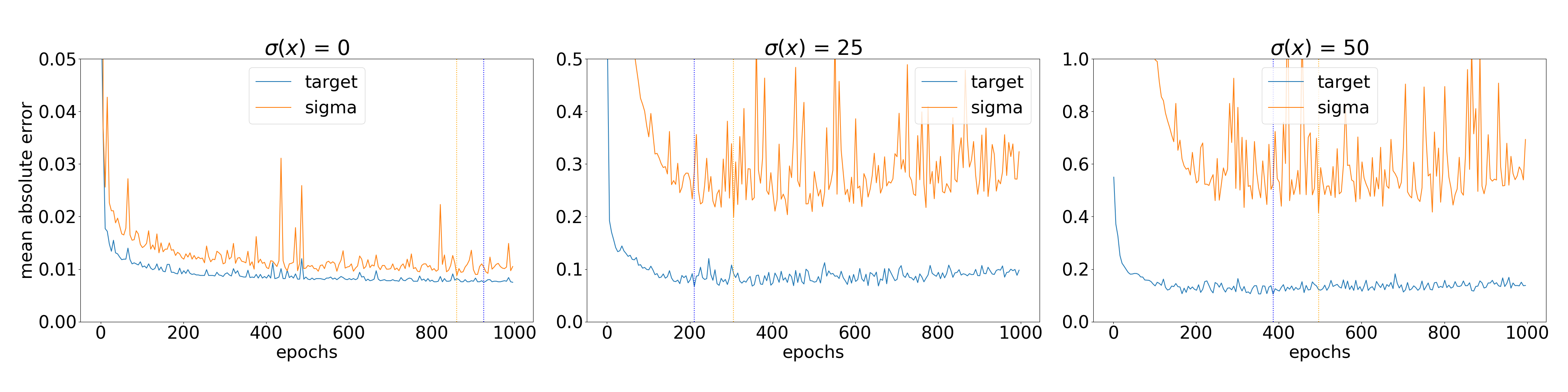}
\caption{\textbf{First row}: test errors in the true target estimate. The stopping epoch suggested by early stopping for the original model is shown by the vertical dotted line. \textbf{Second row}: test errors in the target and sigma estimates by the two-head (2x) model, and best epochs are indicated by the two vertical dotted lines.} \label{fig:target_error}
\end{figure}

The first row of Fig.~\ref{fig:target_error} compares the test errors in the true target estimate by the three models. When the target noise is negligible, all three have relatively stable error curves, and the original model consistently outperforms the others. But when the target noise is high, the original model starts to overfit within 10 epochs, and the accuracy declines faster than the two-heads. This confirms that the uncertainty estimate $\hat{\sigma}^2(x)$ regularizes the training loss, as suggested by \cite{kg}. However, before overfitting, the original model achieved the lowest error among all. Therefore, rather than training it for a predetermined number of epochs, we apply early stopping with the validation MSE criterion. As shown in these plots, early stopping effectively prevents overfitting and leads to better estimates. Further, the second row of Fig. \ref{fig:target_error} shows the test errors in the target and sigma estimates by the two-head (2x) model. As indicated by the two vertical lines, the best epochs are substantially separated, and the estimand with the smaller magnitude gets to the optimal faster. Therefore, even with early stopping, we cannot simultaneously obtain both optimal estimates. As a comparison, we train a second ESPCN with the same NLL loss to estimate only sigma (i.e., the model structure is unchanged). In this loss, the target estimates are obtained using the first (original) ESPCN in the previous experiment. For fairness, we use the same learning rate in the second model, and apply early stopping with the validation NLL criterion to all models. Table \ref{tab:sp_sigmamae} compares the test mean absolute error (MAE) in the sigma estimate, and the original model achieves the best results at all noise levels (each result is computed with 4 random seeds).

We have shown that estimating the true target and aleatoric uncertainty in two steps outperforms the joint model with parameter sharing; in addition, doubling the number of parameters for the two-head model does not have a significant impact on the results.

\begin{table*}[h]
    \centering
    \caption{Test MAE in the sigma estimate ($\star$ means statistically significant).} \label{tab:sp_sigmamae}
    \begin{tabular}{ |l|c|c|c| } 
     \hline    
     $\sigma(x)$ & 0 & 25 & 50 \\ \hline
     Original  & \textbf{0.0085} $\pm$ $0.0001^\star$ & \textbf{0.1896} $\pm$ 0.0114 & \textbf{0.3880} $\pm$ 0.0109 \\ 
     Two-head & 0.0108 $\pm$ 0.0004 & 0.1909 $\pm$ 0.0053 & 0.3983 $\pm$ 0.0375  \\ 
     Two-head (2x) & 0.0101 $\pm$ 0.0002 & 0.1966 $\pm$ 0.0072 & 0.4088 $\pm$ 0.0167  \\     
     \hline
    \end{tabular}
\end{table*}

\subsection{MRI Reconstruction}

We run the MRI reconstruction experiment on the fastMRI dataset, where no synthetic noise is added to the targets. We use the Adaptive-CS-Net (ACSNet) \cite{nicola}, which won the multi-coil track of the 2019 fastMRI challenge\footnote{\url{https://fastmri.org}}. It uses a recurrent neural network to iteratively improve the target estimate in k-space, and its training loss is a combination of the $L_1$ and SSIM losses in the image space. There are two output channels in each iteration, which represent the real and imaginary components of the k-space estimate.  A diagram of the model is shown in Figure 1 of their paper.

We compare with three state-of-the-art predictive uncertainty quantification methods, which are Monte Carlo dropout (MC-D) \cite{mcdropout}, Kendall and Gal (K\&G) \cite{kg} and deep ensembles (DE) \cite{deepensemble}. Since dropout \cite{dropout} is not used in the ACSNet, we follow \cite{fastMRI} and insert a dropout layer after each convolution block. To implement the two-head model, it is not feasible to duplicate the last layer due to the recurrent model structure; therefore, we use the model's last two outputs as the image and uncertainty estimates, and train with the NLL loss. For MC-D and K\&G, we use dropout rates 0.05, 0.1, 0.2 and 0.5, and draw 50 samples for evaluation. For DE, we follow \cite{deepensemble} and train an ensemble of 5 models. We use the 16GB Nvidia Tesla V100 GPUs, and training one model takes 62 hours and 12GB memory. Due to hardware limits, we cannot implement the fine-tuning step in the original paper; in addition, we cannot double the number of parameters for K\&G and DE, or run the adversarial training \cite{adv_training} for DE as it needs to store the gradients twice per input. However, we have run these experiments on a U-net \cite{unet} model\footnote{The original U-net code is from: \url{https://github.com/facebookresearch/fastMRI}.} and found the memory-intensive experiments bring little extra benefits. For our method, we follow \cite{deepensemble} and train an ensemble of 5 models to estimate the target, and 1 model to estimate the absolute error with the $L_1$ loss. For an ablation study, we also train 1 model to estimate the target and 1 model the absolute error (results are marked with ``Ours (one)" in Table \ref{tab:acsnet_mri}). For ease of comparison, we use Adam with learning rate $10^{-4}$ for 40 epochs, $10^{-5}$ for 10 more epochs, and apply early stopping with the validation $L_1$ criterion to all models. Some of the fully sampled test MR images are noisy, but we cannot access the true targets (i.e. \textit{clean} and fully sampled images), so we use these noisy targets as ground truths for target evaluation. We note this makes the error prediction very challenging, since the noise levels greatly vary among test images. We follow \cite{fastMRI} and report the NMSE, PSNR and SSIM results. For  error evaluation, we report the absolute error prediction accuracy in $L_1$ and MSE. Lastly, since baseline methods do not estimate the systematic error, for fairness, we calibrate their uncertainty estimates on the validation set by optimizing: $\alpha_{\star} = \underset{\alpha}{\textbf{min}} \sum_x (|y(x) - \hat{h}(x)| - \alpha \hat{u}(x))^2$, where $\hat{u}(x)$ is the square root of the estimated predictive uncertainty, then the calibrated estimate is $\alpha_{\star}\hat{u}(x)$. 

Table \ref{tab:acsnet_mri} shows our method achieves the best results in all metrics\footnote{The high standard deviations (SD) are due to high image noise (we note that most prior works on this dataset do not report SD, including \cite{fastMRI}).}. Additionally, we have three remarks. First, MC-D estimates only model uncertainty, and the accuracy in error estimates improves as the dropout rate decreases (same for K\&G). This suggests the second random error in Eq. \ref{eq:errordecomp} is very small, which makes sense since we use 50 samples for evaluation. Nevertheless, there is a gap between their results and ours, as we also estimate the systematic error and target noise. Second, K\&G and DE do not perform well in this task, since they modified the training objective and used the last two outputs for image and uncertainty estimates, which changes the dynamics of the model; in addition, they do not estimate the systematic error. Finally, the calibration improves all baseline results, except for MC-D 0.05 (4x). In the supplementary material, we show the uncalibrated baseline results in Table \ref{tab:uncalibrated_results}, and compare the \textit{calibrated} baseline error plots and ours on a random 8x test MRI in Fig. \ref{fig:error_plot_comp}.

\begin{table*}[t]
    \caption{Target and absolute error estimation results for 4x (top) and 8x (bottom) accelerated MRIs. Dropout rates are indicated after ``MC-D" and ``K\&G". Best results are in bold.} \label{tab:acsnet_mri}
    \begin{tabular}{ |l|c|c|c|c|c| } 
     \hline    
     & \multicolumn{3}{|c|}{Target (4x)} & \multicolumn{2}{|c|}{Absolute Error (4x)} \\ \cline{2-6}     
     Method & NMSE($10^{-3}$) $\downarrow$ & PSNR $\uparrow$ &  SSIM $\uparrow$ & $L_1$($10^{-6}$) $\downarrow$ & MSE($10^{-12}$) $\downarrow$ \\ \hline
     MC-D 0.05 & 7.12 $\pm$ 5.97 & 38.46 $\pm$ 2.97 & 0.912 $\pm$ 0.062 & 1.98 $\pm$ 0.92 & 9.62 $\pm$ 14.91 \\
     MC-D $0.1$ & 7.73 $\pm$ 6.14  & 38.07 $\pm$ 2.97 & 0.909 $\pm$ 0.063 & 2.06 $\pm$ 0.89 & 9.90 $\pm$ 13.83 \\ 
     MC-D $0.2$ & 8.28 $\pm$ 6.29 & 37.58 $\pm$ 2.76 & 0.905 $\pm$ 0.061 & 2.25 $\pm$ 0.99 & 11.92 $\pm$ 15.67 \\ 
     MC-D $0.5$ & 16.60 $\pm$ 17.88  & 35.91 $\pm$ 3.83 & 0.888 $\pm$ 0.072 & 2.81 $\pm$ 1.05 & 17.69 $\pm$ 19.84 \\ 
     K\&G 0.05 & 7.38 $\pm$ 5.78 & 38.11 $\pm$ 2.74 & 0.907 $\pm$ 0.063 & 2.51 $\pm$ 0.96 & 14.55 $\pm$ 2.25 \\
     K\&G $0.1$ & 8.69 $\pm$ 6.72 & 37.24 $\pm$ 2.53  & 0.898 $\pm$ 0.063 & 2.90 $\pm$ 1.10 & 19.70 $\pm$ 30.72 \\
     K\&G $0.2$ & 10.47 $\pm$ 7.21 & 36.22 $\pm$ 2.28  & 0.889 $\pm$ 0.062 & 3.39 $\pm$ 1.30 & 27.51 $\pm$ 40.92 \\
     K\&G $0.5$ & 43.13 $\pm$ 11.33 & 29.67 $\pm$ 2.34 & 0.778 $\pm$ 0.070 & 9.85 $\pm$ 4.14 & 174.60 $\pm$ 254.10 \\
     DE & 7.58 $\pm$ 5.41 & 37.99 $\pm$ 2.74 & 0.905 $\pm$ 0.063 & 1.98 $\pm$ 1.01 & 10.36 $\pm$ 17.22 \\ 
     Ours (one) & 6.56 $\pm$ 6.00 & 38.87 $\pm$ 3.03 & 0.915 $\pm$ 0.063 & 1.77 $\pm$ 0.82 & 7.77 $\pm$ 13.10 \\     
     Ours & \textbf{6.34} $\pm$ 5.92 & \textbf{39.08} $\pm$ 3.11 & \textbf{0.917} $\pm$ 0.062 & \textbf{1.74} $\pm$ 0.79 & \textbf{7.32} $\pm$ 12.59 \\ \hline
     & \multicolumn{3}{|c|}{Target (8x)} & \multicolumn{2}{|c|}{Absolute Error (8x)} \\ \cline{2-6}     
     Method & NMSE($10^{-2}$) $\downarrow$ & PSNR $\uparrow$ &  SSIM $\uparrow$ & $L_1$($10^{-6}$) $\downarrow$ & MSE($10^{-11}$) $\downarrow$ \\ \hline
     MC-D 0.05 & 1.23 $\pm$ 0.67 & 35.51 $\pm$ 2.43  & 0.878 $\pm$ 0.064 & 2.74 $\pm$ 1.53  & 2.31 $\pm$ 3.49 \\
     MC-D $0.1$  & 1.32 $\pm$ 0.70  & 35.22 $\pm$ 2.40 & 0.875 $\pm$ 0.063 & 2.84 $\pm$ 1.56  & 2.34 $\pm$ 3.35 \\ 
     MC-D $0.2$ & 1.52 $\pm$ 0.76  & 34.55 $\pm$ 2.38  &  0.867 $\pm$ 0.063 & 3.10 $\pm$ 1.65 & 2.93 $\pm$ 4.02 \\ 
     MC-D $0.5$ & 2.92 $\pm$ 1.96  & 32.09 $\pm$ 2.83 & 0.834 $\pm$ 0.074 & 4.20 $\pm$ 2.26 & 4.91 $\pm$ 6.45 \\
     K\&G 0.05 & 1.54 $\pm$ 0.69 & 34.41 $\pm$ 2.30 & 0.863 $\pm$ 0.064 & 4.50 $\pm$ 1.79 & 5.11 $\pm$ 7.94 \\
     K\&G $0.1$ & 1.91 $\pm$ 0.86  & 33.40 $\pm$ 2.31 & 0.849 $\pm$ 0.062 & 5.24 $\pm$ 2.02 & 6.95 $\pm$ 10.39 \\
     K\&G $0.2$ & 2.23 $\pm$ 0.91  & 32.70 $\pm$ 2.31 & 0.838 $\pm$ 0.061 & 5.81 $\pm$ 2.33 & 8.79 $\pm$ 13.26 \\
     K\&G $0.5$ & 7.37 $\pm$ 2.84  & 27.48 $\pm$ 2.66 & 0.720 $\pm$ 0.074 & 12.82 $\pm$ 5.76 &  35.86 $\pm$ 56.94 \\
     DE  & 1.55 $\pm$ 0.69 & 34.35 $\pm$ 2.30 & 0.861 $\pm$ 0.063 & 3.01 $\pm$ 1.83 & 3.04 $\pm$ 4.52 \\ 
     Ours (one) & 1.21 $\pm$ 0.70 & 35.62 $\pm$ 2.48 & 0.880 $\pm$ 0.065 & 2.54 $\pm$ 1.46 & 2.28 $\pm$ 3.95 \\
     Ours  & \textbf{1.13} $\pm$ 0.68 & \textbf{35.98} $\pm$ 2.52 & \textbf{0.884} $\pm$ 0.065 & \textbf{2.44} $\pm$ 1.38 & \textbf{2.00} $\pm$ 3.44 \\
     \hline
    \end{tabular}
\end{table*}

\section{Related Work}

To assess the predictive quality without ground truth, the reverse classification accuracy framework \cite{valindria}, and a regression algorithm using shape and appearance features \cite{kohlberger} have been proposed, though these methods are limited to the segmentation task. Predictive quality can also be evaluated by the calibrated confidence, which estimates the frequency of the target falling in a given interval \cite{guo,kuleshov,ovadia}. In addition, the generalization error, which measures the prediction accuracy on unseen data, can be estimated by cross-validation, but this is done on a separate test set \cite{hastie}. Further, when the estimands are related to each other, the joint model with parameter sharing can be effective \cite{goodfellow}. Lastly, other interesting uncertainty estimation methods in medical imaging include \cite{adler,kwon,tanno}, etc.

\section{Conclusion}

Current methods in deep learning estimate predictive uncertainty by the sum of data and model uncertainties. In this work, we show this estimate cannot highly correlate with prediction error; in addition, estimating the target and uncertainty separately outperforms the joint model by the current methods. Following this, we propose a novel two-step method that can accurately estimate the target and magnitude of the prediction error on unseen in-distribution data. For future work, we would like to extend this method to tackle out-of-distribution detection. \\

\noindent \textbf{Acknowledgements.} We thank Tony O'Hagan, Yoshua Bengio and the anonymous reviewers for helpful discussions. This research was supported by the NWO Perspective Grant DLMedIA and in-cash and in-kind contributions by Philips.

%
%
%
\bibliographystyle{splncs04}
\bibliography{paper1009}

\appendix
\renewcommand{\thetable}{S\arabic{table}}
\renewcommand{\thefigure}{S\arabic{figure}}
\setcounter{table}{0} 
\setcounter{figure}{0} 

\section*{Supplementary Material}

\begin{table}[h]
    \centering
    \caption{The uncalibrated baseline absolute error estimation results.} 
    \label{tab:uncalibrated_results}
    \begin{tabular}{ |l|c|c| } 
     \hline    
     & \multicolumn{2}{|c|}{Absolute Error (4x)} \\ \cline{2-3}     
     Method & $L_1$($10^{-6}$) $\downarrow$ & MSE($10^{-12}$) $\downarrow$ \\ \hline
     MC-D 0.05 & 2.34 $\pm$ 1.08 & 14.28 $\pm$ 20.39 \\
     MC-D 0.1 & 2.34 $\pm$ 0.99 & 14.41 $\pm$ 21.94 \\
     MC-D 0.2 & 2.34 $\pm$ 1.13  & 14.76 $\pm$ 19.85 \\     
     MC-D 0.5 & 2.36 $\pm$ 1.26  & 15.01 $\pm$ 13.51 \\    
     K\&G 0.05 & (99.96 $\pm$ 0.04) $\times 10^4$ & (99.91 $\pm$ 0.07) $\times 10^{10}$ \\
     K\&G 0.1 & (99.97 $\pm$ 0.03) $\times 10^4$ &  (99.93 $\pm$ 0.04) $\times 10^{10}$  \\     
     K\&G 0.2 & (99.97 $\pm$ 0.06) $\times 10^4$ & (99.93 $\pm$ 0.05) $\times 10^{10}$ \\     
     K\&G 0.5 & (99.99 $\pm$ 0.01) $\times 10^4$ & (99.99 $\pm$ 0.01) $\times 10^{10}$ \\     
     DE & 2.31 $\pm$ 1.36 & 12.27 $\pm$ 9.18 \\
     \hline
     & \multicolumn{2}{|c|}{Absolute Error (8x)}  \\ \cline{2-3}     
     Method & $L_1$($10^{-6}$) $\downarrow$ & MSE($10^{-11}$) $\downarrow$ \\ \hline
     MC-D 0.05 & 2.98 $\pm$ 1.69 & 3.16 $\pm$ 3.73 \\
     MC-D 0.1 & 3.07 $\pm$ 1.68 & 3.23 $\pm$ 4.69 \\
     MC-D 0.2 & 3.62 $\pm$ 1.42 & 4.00 $\pm$ 8.14 \\     
     MC-D 0.5 & 4.26 $\pm$ 2.01 & 5.88 $\pm$ 7.85 \\    
     K\&G 0.05 & (99.96 $\pm$ 0.03) $\times 10^{4}$ & (99.92 $\pm$ 0.07) $\times 10^{9}$ \\
     K\&G 0.1 & (99.97 $\pm$ 0.03) $\times 10^{4}$ & (99.92 $\pm$ 0.06) $\times 10^{9}$ \\     
     K\&G 0.2 & (99.97 $\pm$ 0.03) $\times 10^{4}$ & (99.93 $\pm$ 0.05) $\times 10^{9}$ \\     
     K\&G 0.5 & (99.99 $\pm$ 0.01) $\times 10^{4}$ & (99.98 $\pm$ 0.02) $\times 10^{9}$ \\     
     DE & 3.52 $\pm$ 2.04 & 3.43 $\pm$ 4.72 \\
     \hline     
    \end{tabular}
\end{table}

\begin{figure}[t]
    \centering
    \includegraphics[width=0.9\textwidth]{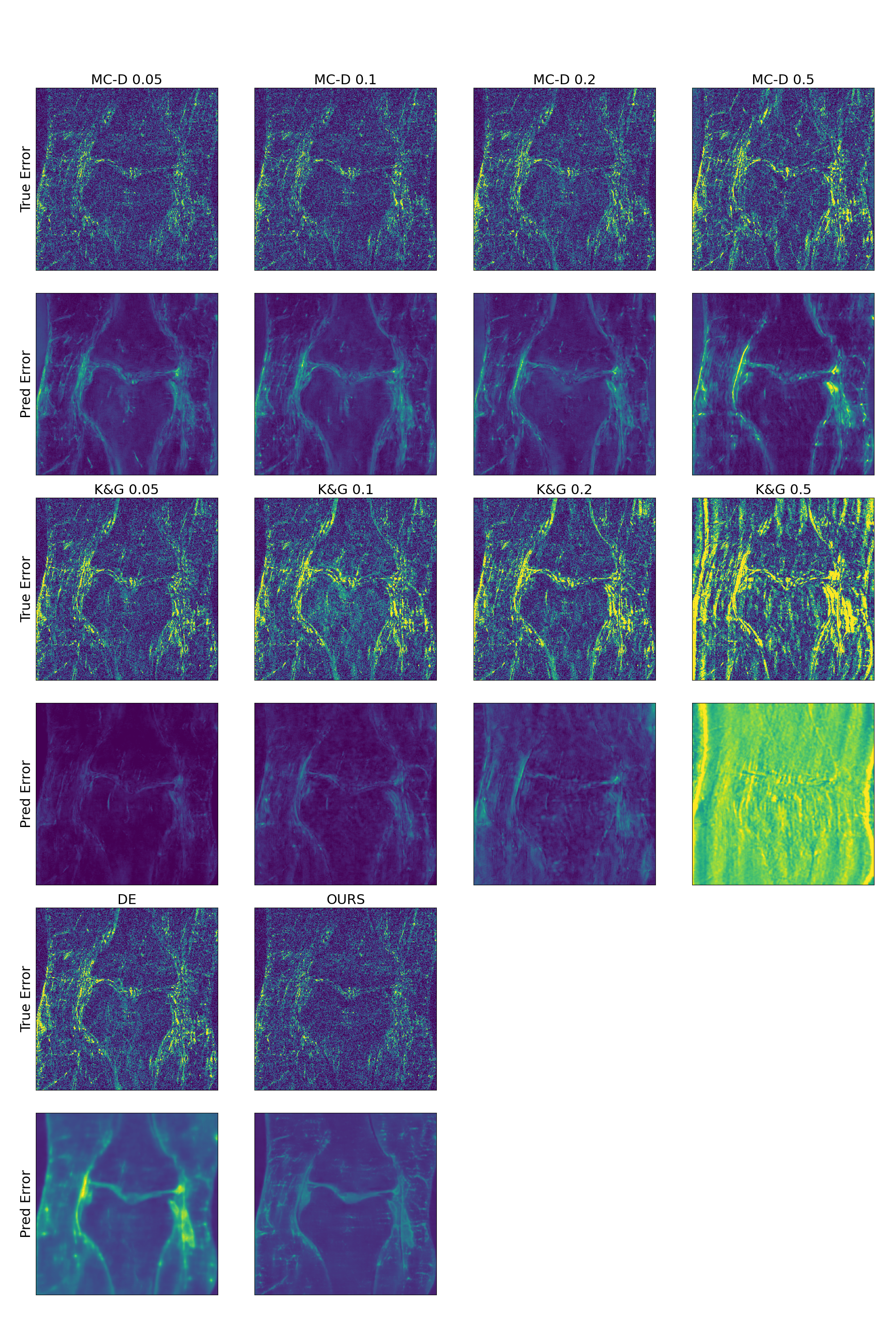}
    \caption{The absolute error predictions on a random 8x test MRI. All plots are clipped to the same pixel range, and all baseline predictions are calibrated. Our prediction shows a clear structure, and has the right pixel intensities in both the foreground and background.}
    \label{fig:error_plot_comp}
\end{figure}

\end{document}